\documentclass[a4paper,11pt]{article}
\pdfoutput=1 

\usepackage{jheppub} 
\usepackage{graphicx}
\usepackage{caption}
\usepackage{subcaption}
\usepackage[T1]{fontenc} 
\graphicspath{{plots/}}

\title{\boldmath A Loss-Function for Causal Machine-Learning}

\author[a]{I-Sheng Yang,}

\affiliation[a]{Independent Researcher}

\emailAdd{isheng.yang@gmail.com}

\abstract{Causal machine-learning is about predicting the net-effect (true-lift) of treatments.
Given the data of a treatment group and a control group, it is similar to a standard supervised-learning problem.
Unfortunately, there is no similarly well-defined loss function due to the lack of point-wise true values in the data.
Many advances in modern machine-learning are not directly applicable due to the absence of such loss function.

We propose a novel method to define a loss function in this context, which is equal to mean-square-error (MSE) in a standard regression problem.
Our loss function is universally applicable, thus providing a general standard to evaluate the quality of any model/strategy that predicts the true-lift.
We demonstrate that despite its novel definition, one can still perform gradient descent directly on this loss function to find the best fit.
This leads to a new way to train any parameter-based model, such as deep neural networks, to solve causal machine-learning problems without going through the meta-learner strategy.
}

\begin{document} 
\maketitle
\flushbottom

\section{Introduction}

Causal machine-learning with experimental data is also known as the true-lift problem, the uplift problem, the net-lift problem, or the conditional average treatment effect (CATE).
The number of different names comes from the fact that it is a very general problem found across many fields like medicine, economics, and political science \cite{GreKer2012, TiaEtal2014, AthImb2015, WyrJar2015, HenEtal2016, TadEtal2016, PowEtal2018, GubSte2019}.
This book \cite{MicIgo2019Book} provides a good introduction, status summary, and links to literature.

The section 2 of this recent paper \cite{KunEtal2018} provides a rigorous mathematical definition following the framework of Rubin \cite{Rubin1974}.
Here, we will start by visualizing one of the core technical challenges in causal machine-learning, that is to deal with training data in the following format. 
\begin{center}
\begin{tabular}{ |c|c|c|c| } 
 \hline
 Features & Value with Treatment & Value without Treatment 
 & $l = y^T - y^C$ \\ 
 \hline
$x_1$ & $y_1$ & ? & ? \\ 
 \hline
 $x_2$ & $y_2$ & ? & ? \\ 
 \hline
 $x_3$ & $y_3$ & ? & ? \\ 
 \hline
 $x_4$ & ? & $y_4$ & ? \\ 
 \hline
 $x_5$ & ? & $y_5$ & ? \\ 
 \hline
\end{tabular}
\end{center}
Every row represents a case within a mass experiment.
For example, in the medical context it can be about a treatment on a patient, with the ``Features'' as the medical history and the ``Value'' as an index of health after the experiment.
In the marketing context, it can be about a promotion sent to a consumer, with ``Features'' as the shopping behavior and demographic information, and the ``Value'' as the money spent during the promotional period.

The experiment was carefully conducted to enable an A/B test.
With a random separation, we have a treatment group where the treatment was actually administered, and a control group where the treatment was not administered (or administered a carefully prepared placebo instead).
The goal is to predict the net effect of the treatment, also known as the ``Lift'' or ``True-Lift'', that is defined as the value $y$ if the treatment was given, subtract the value $y$ if the treatment was not given, abbreviated as $l = y^T$ (treatment) $-y^C$ (control).

If we think within the framework of supervised learning, we are trying to search within a family of functions that return predictions
\begin{equation}
    f(\lambda, x_i) = p_i~,
\end{equation}
and to find the parameter $\hat\lambda$ which minimizes the loss function on the training set,
\begin{equation}
{\rm MSE} = \frac{1}{\|\{{\rm training \ set}\}\|}\sum_i (p_i - l_i)^2~.
\label{eq-loss}
\end{equation}
This is unfortunately not doable, because the loss function in Eq.~(\ref{eq-loss}) cannot be defined.
The definition of Eq.~(\ref{eq-loss}) requires the point-wise true values, $l_i$, which we are trying to predict.
However, it is obvious from the above table that every single case can only be in either the treatment group or the control group, but never both.
Therefore, the point-wise lift value $l_i$ is never actually known.

The main research to address this problem can be viewed as a collection of different efforts to go around the unfortunate fact that the loss function in Eq.~(\ref{eq-loss}) cannot be defined.
Roughly speaking, most these efforts fall into one of the following 2 categories.\footnote{
This github source (https://github.com/uber/causalml) contains the implementation of models in both categories.
}

The first category has a recently coined name---meta-learners \cite{KunSek2019, NieWag2017}.
The basic idea is probably the first solution ever proposed.
Instead of predicting the lift value directly, a meta-learner is an indirect strategy that goes through some intermediate predictions.
Of course, most importantly, it is usually based on the predictions of the treatment response $y^T$ and the control response $y^C$.
These predictions are standard supervised-learning problems and can be done in various ways.
After that, we can take the difference of the two predictions as the predicted lift.
There are also a few more elaborated strategies based on the predictions of $y^T$ and $y^C$, and this type of strategies all fall into this category.

The advantage of meta-learners is in the flexibility.
It only prescribes a meta-strategy on how to combine models. 
The actual machine-learning models employed within the strategy can be anything we want. 
On the other hand, this comes with a very real disadvantage.
Whenever we train/fit a model, it is an optimization process.
We all know that optimization does not commute well with other operations.
In a meta-learner, every optimization that happens as an intermediate step has no reason to be exactly aligned with the final goal of getting the best prediction of $l$.\footnote{
In fact, one can find counter-examples---less optimized predictions of $y^T$ and $y^C$ can sometimes lead to better final predictions of $l$.
It depends on how the errors of intermediate predictions are correlated, which is not a factor considered while optimizing for those predictions.}

The other category contains tree-based models \cite{RzeJar2012, SolRze2015}.
Even in a standard supervised learning problem, a decision tree does not try to directly minimize a final loss function.
Instead, it operates on a loss function that is defined for the purpose of splitting a node into 2 sub-nodes.
Combine this split-loss-function with smart strategies to search through different ways of splitting the node, the resulting tree which grew from a series of the optimized splittings can be quite aligned with optimizing the final loss function.

The advantage here is that, unlike the final loss function, the split-loss-function is directly generalizable to an A/B test training data.
In order words, for any known strategy of building a decision tree in a standard supervised learning problem, we have a corresponding strategy to build a decision tree for the true-lift problem.
It is directly optimizing for the lift prediction, in exactly the same sense as using the original decision tree in a standard supervised learning problem.

This is in sharp contrast to the meta-learner strategies.
Incidentally, the disadvantage here is also exactly on the flip side.
Tree-based models are somewhat limited.
It includes decision trees, ensemble of trees like random forest, and basically nothing else.\footnote{
Not even boosted trees are included, since the idea of boosting requires the point-wise true values.}
Nothing in the deep learning framework are directly compatible with tree-based models.
This is really unfortunate.
Deep learning is one of the most popular branch of machine learning at this moment.
If the features come in the form of an image, natural language, or a relational graph, there are go-to deep learning models one really wants to use.

As it stands, machine-learning practitioners can run into such a dilemma.
Given a true-lift problem with input data in the form of images, text or graphs, meta-learners are very convenient.
We can take advantage of the well-established, sometimes even pre-trained deep-learning models, and assemble the meta-learners with minimal effort.
However, we will have to live with the fear that optimization of these models does not directly translate into the optimization of the lift prediction.
Alternatively, we can insist on using a tree-based model that is optimized for predicting the true-lift.
Nevertheless, it will be a very daunting effort of feature engineering to put the data into an acceptable format for the tree-based model.

In this paper, we will provide a third way out.
We will first go back to the loss function in Eq.~(\ref{eq-loss}) and show that {\bf it is actually definable even without point-wise true values.} 
If the predictions take discrete values, our definition can be derived directly from Eq.~(\ref{eq-loss}).
For continuous predictions, we can first put the values into discrete bins and then treat them as discrete values.

Some might consider this binning as a loss of generality.
We should point out that it is actually necessary.
The quality of true-lift predictions can never be defined unless we assume certain grouping structure. 
Such structure determines how subsets of treatment and control data are combined to compute the lift values.
If we think about why tree-based models work naturally, it is because the nodes within a tree provide a natural way to determine such grouping structure.
Unfortunately, the nodes are intrinsic to trees and do not generalize to other types of models.
In order to have a universal way to quantify the quality of true-lift predictions, the grouping structure must be based on quantities that always exist.
It should be independent of a model's internal architecture, parameters, intermediate variables, and also independent of which features are used.
Since the only universal requirement for a true-lift model is to output the prediction values, binning them for grouping is the only reasonable solution.
With the freedom to choose the number of bins, the loss of generality is minimal. 
Our loss function is universally applicable to any true-lift predictions, and provides a general standard to evaluate the performance of any true-lift models.

Furthermore, we will provide a full recipe to perform gradient descent directly on such loss function.
Starting from a continuous model, we will go through the binning process, and then a few technical steps to calculate the gradient on our loss function.
During gradient descent, a model will simultaneously learn to give more accurate predictions within each group, and also learn to evolve the grouping structure for better predictions across the groups.
Although there are some extra hoops to jump through, the entire algorithm can still run efficiently on big data.
Our loss function and recipe for gradient descent allow any parameter-based model to be trained on A/B-test data and directly optimized for the lift prediction.
It opens up the 3rd avenue for solving causal machine-learning problems.
In the particular case of causal deep learning, this 3rd avenue seems to be a good direction to go.

\subsection{Outline}

In Section \ref{sec-dis_loss}, we will establish the definition of the loss function for models that provide discrete predictions, and prove that it is essentially the same as MSE.
In Section \ref{sec-gradient}, we will demonstrate that although the definition of gradient is slightly more involved with this loss function, it is still definable and can be efficiently computed on big data.
We will use a simple example to visualize how it works almost the same as vanilla gradient descent, and is indeed improving the quality of the true-lift predictions.
In  Section \ref{sec-dis}, we make some concluding remarks and also point out a few interesting future directions.

\section{Loss Function for Discrete Models}
\label{sec-dis_loss}

A model that provides discrete predictions is basically a function $f(\lambda, x)$ such that
\begin{equation}
    \forall x~, \ \ f(\lambda, x)=p~\in~\{P_n|n=1,2,...N\}~.
\end{equation}
Of course, while training the model, we will assume that both the treatment set $S^T$ and the control set $S^C$ in the training data are dense enough to populate all possible prediction values.
\begin{eqnarray}
S^T_n \equiv \{x_i| f(\lambda, x_i)=P_n, x_i\in S^T\}~,
& &
S^C_n \equiv \{x_i| f(\lambda, x_i)=P_n, x_i\in S^C\}~,
\nonumber \\
\|S^T_n\|\gg1~,& & \|S^C_n\|\gg1~, \ \ \forall~n~.
\end{eqnarray}
We will adopt the notation that the total training data is $S \equiv S^T\cup S^C$, and this relation generalizes to the subset defined by the prediction values, $S_n\equiv S^T_n\cup S^C_n$.
Finally, we will assume that the effective random separation between treatment control extends to the subsets.
\begin{equation}
    \frac{\|S_n\|}{\|S\|}\approx
    \frac{\|S^T_n\|}{\|S^T\|}\approx
    \frac{\|S^C_n\|}{\|S^C\|}~,
\end{equation}
where the norm $\|S\|$ means the size of a set (number of rows).\footnote{
This basically means that the separation between treatment and control is entirely random and infinitely stratified on all of the features $x$.
Namely, it is impossible to use any information from the features $x$ to tell whether a point is more likely to be in the treatment set or the control set.
Of course, when the training set is finite in size, this can only be approximately true.
}

Given this setup, we will revisit the loss function in Eq.~(\ref{eq-loss}), and we can start by rewriting the summation over all data points as a two-stage summation.
\begin{eqnarray}
{\rm MSE}(\lambda) &=& \frac{1}{\|S\|} \sum_{n=1}^N
\sum_{x_i\in S_n} 
\left[ f(\lambda, x_i) - l_i \right]^2
\label{eq-remove_cross}
\\ \nonumber
&=& \frac{1}{\|S\|} \sum_{n=1}^N
\sum_{x_i\in S_n} 
\left[ f(\lambda, x_i) -\bar{l}_n + \bar{l}_n - l_i \right]^2
\\ \nonumber
&=& \frac{1}{\|S\|} \sum_{n=1}^N
\bigg[ \|S_n\| (P_n-\bar{l}_n)^2 
+ \sum_{x_i\in S_n} (\bar{l}_n-l_i)^2 \bigg]~.
\end{eqnarray}
Here we introduced the quantity $\bar{l}_n$ that is the average lift within a subset.
\begin{equation}
    \bar{l}_n \equiv \frac{1}{\|S_n\|}
    \sum_{x_i\in S_n} l_i
    \approx\bar{y}_n^T-\bar{y}_n^C
    = 
    \bigg(
    \frac{1}{\|S^T_n\|}
    \sum_{x_i\in S^T_n} y_i
    -\frac{1}{\|S^C_n\|}
    \sum_{x_i\in S^C_n} y_i\bigg)~.
\end{equation}
There are clear advantages for introducing this quantity.
First of all, by its definition, the cross term in Eq.~(\ref{eq-remove_cross}) vanishes, so we can separate the loss function into 2 square terms.
Secondly, the subset-average lift can be computed from the actually available training data, without using the point-wise values of $l_i$ which are never really known.

At this point, we are half-way to our final goal.
The loss function is rewritten as the sum of two terms, and one of them can already be defined without the point-wise true values of $l_i$.
The other half of our job is to invoke the well-known fact in statistical analysis: Total variance is the sum of variance explained and the remaining variance.
\begin{eqnarray}
\frac{1}{\|S\|} \sum_i (l_i - \bar{l})^2
&=& \frac{1}{\|S\|} \sum_{n=1}^N
\sum_{x_i\in S_n} 
(l_i - \bar{l}_n + \bar{l}_n - \bar{l})^2
\label{eq-variance}
\\ \nonumber
&=& \frac{1}{\|S\|} \sum_{n=1}^N
\bigg[
\sum_{x_i\in S_n} (l_i - \bar{l}_n)^2
+ \|S_n\| (\bar{l}_n-\bar{l})^2
\bigg]~.
\end{eqnarray}
Naturally, the global average lift $\bar{l}$ is another quantity that we can define without the point-wise true values of $l_i$.
\begin{equation}
    \bar{l} \equiv \frac{1}{\|S\|}
    \sum l_i
    \approx
    \bar{y}^T - \bar{y}^C
    =\bigg(
    \frac{1}{\|S^T\|}
    \sum_{x_i\in S^T} y_i
    -\frac{1}{\|S^C\|}
    \sum_{x_i\in S^C} y_i\bigg)~.
\end{equation}

Combine Eq.~(\ref{eq-remove_cross}) and Eq.~(\ref{eq-variance}), we get
\begin{eqnarray}
{\rm MSE}(\lambda) &=&
\sum_{n=1}^N \frac{\|S_n\|}{\|S\|} 
\bigg[(P_n-\bar{l}_n)^2 - (\bar{l}_n-\bar{l})^2
\bigg]
+\frac{1}{\|S\|}\sum_i (l_i-\bar{l})^2~.
\label{eq-combined}
\end{eqnarray}
Although the last term in Eq.~(\ref{eq-combined}) still requires the point-wise true values of $l_i$, this term has no dependence on the model parameter $\lambda$.
Therefore, while minimizing the loss function, it is basically a constant that can always be ignored.

{\bf At this point, we are ready to present the loss function for the true-lift problem.}
\begin{equation}
    L(\lambda) = \sum_{n=1}^N \frac{\|S_n\|}{\|S\|} 
\bigg[(P_n-\bar{l}_n)^2 - (\bar{l}_n-\bar{l})^2
\bigg]~.
\label{eq-TLloss}
\end{equation}
Its physical meaning is also very clear.
The first term, $(P_n-\bar{l}_n)^2$ is the bias. 
Ideally, we want the predicted value $P_n$ for the subset $S_n$ to agree with the average lift within $S_n$.
A better agreement here naturally means a better model.
On the other hand, a model can still be terrible even with zero bias.
For example, if the average lift $\bar{l}_n$ of every subset is just equal to the global average lift $\bar{l}$, then the model actually predicts nothing.
The second term in Eq.~(\ref{eq-TLloss}) represents the model's capability to separate the training data into subsets whose $\bar{l}_n$ are very different from one another.
Therefore, it makes perfect sense that the combination of these 2 terms is exactly the loss function.\footnote{
One choice of the split-loss-function in a tree-based model can be viewed as a special case of this, where $N=2$ and the bias is by-definition zero.
}

\section{Gradient Descent}
\label{sec-gradient}

\subsection{Definition of Gradient}
\label{sec-def_grad}

Mathematically speaking, it is straightforward to calculate the gradient.
\begin{equation}
    \frac{\partial L}{\partial\lambda} = 
    \frac{L(\lambda + \Delta\lambda)-L(\lambda)}{\Delta\lambda}~.
\end{equation}
However, in the context of machine learning and big data, it is often impractical to calculate the gradient this way.
Evaluating $L$ means going through all the data points in the training set, and the parameter $\lambda$ can be a very high-dimensional vector.
Computing the gradient this way requires one to go through the entire data set too many times.

In practice, the gradient is always calculated through the chain-rule instead.
\begin{equation}
    \frac{\partial L}{\partial\lambda} = 
    \sum_i \frac{\partial f(\lambda, x_i)}{\partial\lambda}
    \frac{\partial L}{\partial p_i}~.
\end{equation}
The first factor, $\partial f(\lambda, x_i)/\partial\lambda$, is given by the internal structure of the model.
Only the second factor cares about the loss function.
In the case of mean-square-error, it is simply
\begin{equation}
    \frac{\partial {\rm MSE} }{\partial p_i} = \frac{2(p_i-l_i)}{\|S\|}~.
\end{equation}
All it needs is to go through the data once to read the true values $l_i$, and then again to compute the predictions $p_i$.
In fact, one is not even required to use all the data at once, but can perform gradient updates in batches if needed.

Therefore, here we will show that with our slightly more involved loss function in Eq.~(\ref{eq-TLloss}), the partial derivative with respect to $p_i$ can still be evaluated efficiently.
We will actually lay out a full procedure of defining such derivative, starting from a model that gives continuous predictions.
\begin{equation}
    f_c(\lambda, x_i) = p_i~.
\end{equation}

First, we need to discretize its predictions in order to define the loss function.
The discretized version of the model can be expressed as
\begin{eqnarray}
f(\lambda, x_i) &=& P_n \equiv 
\frac{\sum_{x_i\in S_n} p_i}{\|S_n\|}~,
\label{eq-discrete}
\\ \nonumber
x_i\in S_n \ \ &{\rm if}& \ \ C_{n-1}<p_i<C_{n}~,
\\ \nonumber
x_i\in S_1 \ \ &{\rm if}& \ \ p_i<C_1~,
\\ \nonumber
x_i\in S_N \ \ &{\rm if}& \ \ p_i>C_{N-1}~.
\end{eqnarray}
We recommend the cut values $C_n$ to be chosen such that the subsets are roughly equal in size.
Note that this only has to be approximately true, therefore we do not need to sort the entire training set to determine $C_n$.

Combine Eq.~(\ref{eq-TLloss}) and (\ref{eq-discrete}), at the first glance, $\partial L/\partial p_i$ seems to be straightforward to compute, since there is only one explicit dependence on $p_i$.
\begin{eqnarray}
    \frac{\partial L}{\partial p_i}\bigg|_{\rm bias}
    &=& \sum_n \frac{\partial L}{\partial P_n}
             \frac{\partial P_n}{\partial p_i}
\\ \nonumber
    &=&  \frac{2}{\|S\|} (P_n-\bar{l}_n)~, \ \ \ x_i\in S_n~.
\label{eq-GradBias}
\end{eqnarray}
This clearly cannot be the full story, since we are only hitting the bias term in Eq.~(\ref{eq-TLloss}).
Updating the model according to the above gradient will only improve the bias term, but it will not improve our ability to separate the data into a more appropriate groups of subsets, which is quite essential for improving the true-lift prediction.

The value of $L$ changes suddenly whenever a data point moves from one subset to another.
That makes $L$ discontinuous, therefore non-differentiable.
Fortunately, gradient descent is just a mean to an end.
The actual goal is to follow the slope and find the minimum of $L$.
For a discontinuous function, we can instead follow the effective gradient that is the average slope across the discontinuity. 
In other words, we should consider a finite change $\Delta p_i$ and how it implicitly changes the value of $L$ through moving the point $x_i$ from a subset $S_n$ into a neighboring one.
\begin{equation}
    \frac{\partial L}{\partial p_i}
    \equiv \frac{\Delta L}{\Delta p_i}
    \equiv \frac{\partial L}{\partial p_i}\bigg|_{\rm bias}
    +\frac{1}{\Delta p_i} \sum_n
    \bigg(
    \frac{\partial L}{\partial \bar{l}_n}\Delta\bar{l}_n
   +\frac{\partial L}{\partial \|S_n\|}\Delta\|S_n\| 
    \bigg)~.
    \label{eq-GradFull}
\end{equation}

In order to do so, we will establish more cut values.
\begin{eqnarray}
    C_n^{\pm} &=& \frac{2}{3}C_n + \frac{1}{3}C_{n\pm1}~,
    \ \ {\rm except} 
    \\ \nonumber
    C_1^- &=& C_1 - \frac{C_2-C_1}{3}~, \\ \nonumber
    C_{N-1}^+ &=& C_{N-1} + \frac{C_{N-1}-C_{N-1}}{3}~.
\end{eqnarray}
Basically, every subset will be further divided into 3 segments.
For the data points within a middle segment, 
\begin{equation}
C_{n-1}^+<p_i<C_{n}^{-}~,
\end{equation}
since they are far away from the discontinuities, we will not consider the possibility of migration and simply use the gradient of the bias term.
Namely, we will keep only the first term in Eq.~(\ref{eq-GradBias}).

For the top segment, 
\begin{equation}
C_{n}^-<p_i<C_{n}~,
\end{equation}
we will compute Eq.~(\ref{eq-GradFull}) assuming that the point $x_i$ migrates upward from $S_n$ to $S_{n+1}$.
For $\Delta p_i$, we will use the following value.
\begin{equation}
\Delta p_i =\frac{C_n-C_n^-}{2}~.
\end{equation}
Note that this is somewhat a free parameter we can choose.
About half of the data points within the top segment will leave the current subset if we shift $p_i$ by $\Delta p_i$.\footnote{
It is likely but not guaranteed to end up in $S_{n+1}$, but that does not matter too much.}
So it is an easily computable value and good enough to estimate the effect.

For $\Delta \|S_n\|$, it is straightforward to use
\begin{equation}
    \Delta\|S_n\| = -1~, \ \ 
    \Delta\|S_{n+1}\| = 1~,
\end{equation}
and zero for all other subsets.
Finally for $\Delta\bar{l}_n$, it depends on whether the point moving is in the treatment group or the control group.
\begin{eqnarray}
{\rm For} \ \ x_i\in S_n^T~, & &
\Delta\bar{l}_n = \frac{-y_i+\bar{y}^T_n}{\|S_n^T\|}~,
\\ \nonumber & &
\Delta\bar{l}_{n+1} = \frac{y_i-\bar{y}^T_{n+1}}{\|S_{n+1}^T\|}~;
\\ \nonumber
{\rm for} \ \ x_i\in S_n^C~, & &
\Delta\bar{l}_n = \frac{y_i-\bar{y}^C_n}{\|S_n^C\|}~,
\\ \nonumber & &
\Delta\bar{l}_{n+1} = \frac{-y_i+\bar{y}^C_{n+1}}{\|S_{n+1}^C\|}~;
\end{eqnarray}
and zero for all other subsets.

For the bottom segment, 
\begin{equation}
C_{n-1}<p_i<C_{n-1}^+~,
\end{equation}
we follow exactly the same logic.
\begin{eqnarray}
\Delta p_i &=& \frac{C_{n-1}-C_{n-1}^+}{2}~, 
\\
\Delta\|S_n\| = -1~, & & 
    \Delta\|S_{n-1}\| = 1~,
\\
{\rm For} \ \ x_i\in S_n^T~, & &
\Delta\bar{l}_n = \frac{-y_i+\bar{y}^T_n}{\|S_n^T\|}~,
\\ \nonumber & &
\Delta\bar{l}_{n-1} = \frac{y_i-\bar{y}^T_{n-1}}{\|S_{n-1}^T\|}~;
\\ \nonumber
{\rm for} \ \ x_i\in S_n^C~, & &
\Delta\bar{l}_n = \frac{y_i-\bar{y}^C_n}{\|S_n^C\|}~,
\\ \nonumber & &
\Delta\bar{l}_{n-1} = \frac{-y_i+\bar{y}^C_{n-1}}{\|S_{n-1}^C\|}~.
\end{eqnarray}

\subsection{Pseudocode and Computation Time}

Given the training set $S$ (or a minibatch), let us summarize the calculation gradient in the form of a step-by-step pseudocode.
\begin{enumerate}
    \item Read the values $y_i$, treatment/control labels, and calculate predictions $f(x_i,\lambda)=p_i$.
    \item Calculate the global lift $\bar{l}$.
    \item Choose the number of subsets $N$. Establish the cut values $C_n$ which separates the subsets.
    \item Compare the values of $p_i$ to $C_n$ to generate the subset label $n$.
    \item Use the subset label and treatment/control label to compute $P_n$, $\bar{y}_n^T$, $\bar{y}_n^C$, $\bar{l}_n$.
    \item Calculate the values of $C_n^{\pm}$ which cut each subset into 3 segments.
    \item Compare the values of $p_i$ to $C_n^\pm$ to generate the segment label top/middle/bottom.
    \item Use the subset label and the segment label to put all of the above values together into Eq.~(\ref{eq-GradFull}) to calculate the gradient.
\end{enumerate}

Let us try to compare the computation time to the usual gradient descent in a regression problem.
Step 1 is always required, and after that, the usual gradient descent needs to perform a tensor subtraction, aa multiplication with $\partial f(\lambda, x_i)/\partial\lambda$, and a summation over $i$.
Here, step 2 is one constant value from the data and does not need to be repeated when we update the gradient.
Steps 3 and 6 do not need to be repeated with every step of the gradient update.
We can set a fixed number of steps before redoing them.
And even when we redo them, we do not need to go through all the data.
So they do not contribute to the computation time either.

The real extra computations here are just a few value comparisons to generate labels which will be used for logical gates, and a few more summations to calculate the subset average values.
Each of these operations takes about the same time as tensor subtraction, multiplication, and summation.
And we should note that in a complicated model, evaluating $p_i$ already involves a big number of tensor operations.
So the fact that we need to perform a few more of those should not pose any problem in practice.

\subsection{Example}

We generated 10k rows of data with 70-30 treatment-control split and the following rules.
\begin{eqnarray}
y^T &=& r_1 + r_2 + 0.5*r_3~, \\
y^C &=& r_1 + r_2~,
\end{eqnarray}
where $r_1$, $r_2$ and $r_3$ are three random numbers.
The visible features are $x= (r_1, r_3)$, and $r_2$ is basically the hidden noise.
Nevertheless, the lift is simply $0.5*r_3$ without any uncertainty.
The information in this data is simple enough that many existing techniques can find the best model.
The purpose here is not comparing different models, but simply as an example to demonstrate and visualize that our gradient descent proposal is working as designed.

\begin{figure}[t]
\begin{subfigure}{.5\textwidth}
  \centering
  \includegraphics[width=1.0\linewidth]{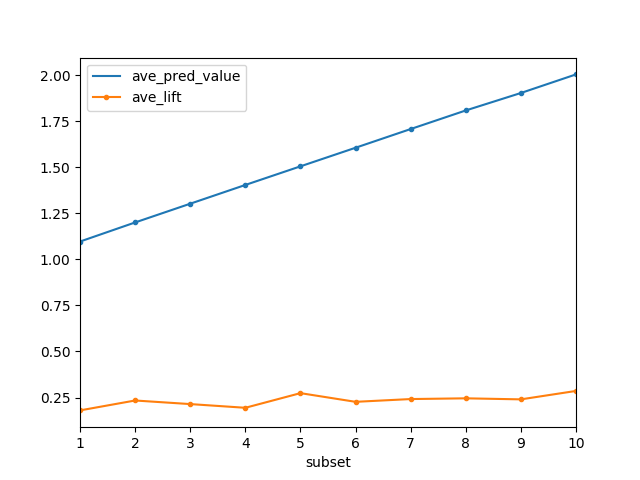}  
  \caption{$t=0$}
  \label{fig:sub-first}
\end{subfigure}
\begin{subfigure}{.5\textwidth}
  \centering
  \includegraphics[width=1.0\linewidth]{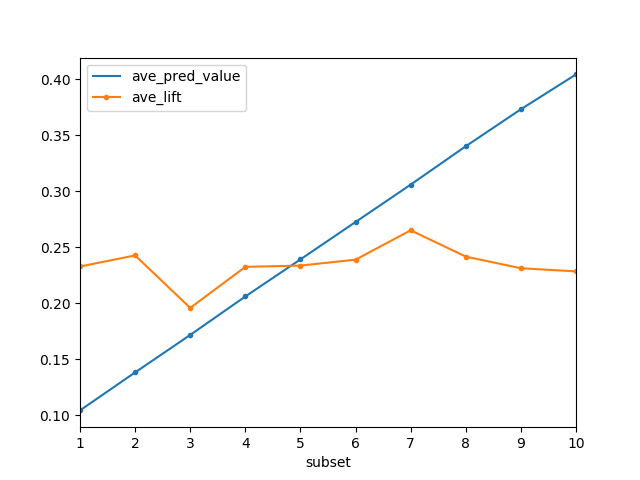}  
  \caption{$t=1$}
  \label{fig:sub-second}
\end{subfigure}
\newline

\begin{subfigure}{.5\textwidth}
  \centering
  \includegraphics[width=1.0\linewidth]{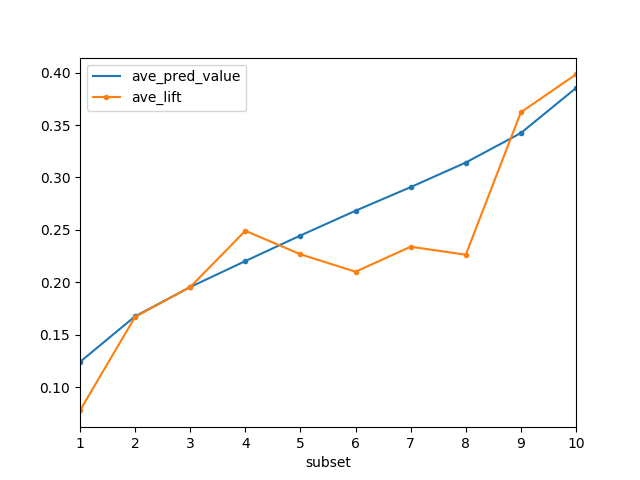}  
  \caption{$t\sim10$}
  \label{fig:sub-third}
\end{subfigure}
\begin{subfigure}{.5\textwidth}
  \centering
  \includegraphics[width=1.0\linewidth]{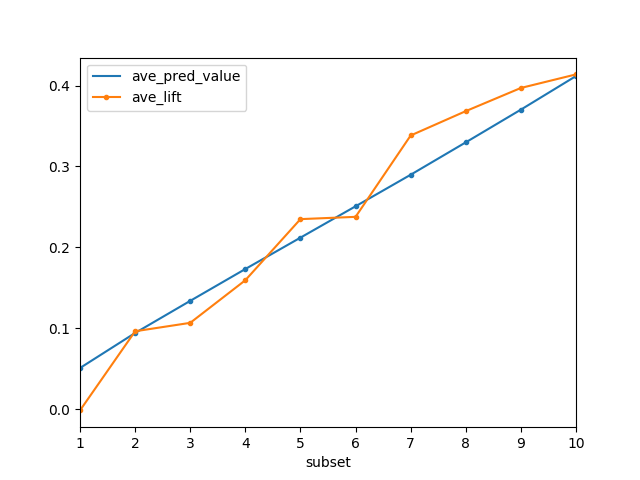}  
  \caption{$t\sim100$}
  \label{fig:sub-fourth}
\end{subfigure}
\caption{Visualizing the average prediction value and average lift of each subset during gradient descent.
We can see that at $t=1$, the global average of predictions is immediately brought down to align with the global average lift.
This is the most obvious way to reduce the bias term in the loss function.
At this time, the separation of subsets is clearly still terrible.
At $t\sim10$, we can see significant migrations between subsets.
The subset-average lifts start to spread out into the range between $0$ to $0.5$.
The top and bottom subsets seem to stand out first, while middle subsets still struggle.
Eventually at $t\sim100$, we can see that the subset-average lifts are aligned nicely with the prediction values span the range between $0$ and $0.5$ as the ideal answer should.
}
\label{fig-GD_process}
\end{figure}

We start with a linear model with offset.
\begin{equation}
    p = c_1*r_1 + c_3*r_3 + c_2~.
\end{equation}
The initial parameter values, $(c_1, c_2, c_3)=(1,1,0.1)$, are deliberately chosen to be quite far away from the true values, $(0,0,0.5)$.
We follow our definition of $\partial L / \partial p_i$ in Sec.\ref{sec-def_grad} and update the linear model with gradient descent with step size $0.1$.
Within 100 steps, our recipe indeed brought the model down to $(c_1, c_2, c_3)=(0.064, -0.001, 0.467)$, which is pretty close to the true values.
The behavior of loss function during this process is visualized in Figure \ref{fig-GD_process}.

We would like to highlight the observation that the process feels exactly the same as a normal gradient descent.
It even runs into similar problems.
For example, in our case, we are not yet bounded away by noises from the true values $(0,0,0.5)$.
We can continue to make small but steady progress toward that.
Unfortunately, we cannot increase step size to make it faster.
That is because the second derivative along the $c_2$ direction is very large.
A bigger step size would have led to an uncontrolled oscillation of $c_2$.
This is the standard narrow-valley problem of first-order gradient descent.
\footnote{Strictly speaking, since we our model really has only 2 orthogonal dimensions, we could have rescaled the $c_2$ axis by a constant to solve this problem.
However, in a higher dimensional model, a curved narrow valley is a real problem that requires more advanced techniques to solve.}

\section{Conclusion and Discussion}
\label{sec-dis}

In this paper, we establish a novel definition of a loss function for predicting true-lift on A/B-test data, and provide a recipe for gradient descent.
Our key insight is that since a certain grouping on such data is always required to define the prediction targets, the only universal approach is to use the model prediction values to define such grouping.
During gradient descent, the model will simultaneously learn to give better predictions on the current groups, and also learn to evolve the grouping structure to increase the lift differences between groups. 
This, for the first time, allow non-tree-based models to be trained directly to predict lifts.
Up to the choice of the number of groups, our loss function is unique and can serve as a general standard to evaluate the performance of any lift predictions.

On the other hand, the recipe for gradient descent we provide is not unique.
That is because, fundamentally, causal problems are deeper than standard supervised problems.
We establish a framework that is highly analogous to standard supervised-learning, but this does not erase the intrinsic, extra complications in causal problems.
All we have done, is to condense and translate those extra complications, into concrete equations and parameters.
It creates a new playing ground for researchers to adjust those parameters and discover better ways to use the equations.
Hopefully, those further efforts will really lead us to dive deeper into the causal machine-learning problem.

Based on this framework, we can already see a few interesting directions to explore further.
\begin{enumerate}
    \item {\bf The number of subsets $N$.} 
    Even if we have determined that in the end, the performance will be evaluated by a fixed value $N$, it does not mean that we have to stick to the same number during gradient descent.
    In fact, it is conceivable that starting with a small number of subsets might make the gradient descent process more efficient.
    And in theory, the best lift prediction is achieved with $N$ as large as possible before statistical error starts to matter.
    Thus tuning $N$ during gradient descent may have surprising effects.
    \item {\bf The value of $\Delta p_i$.}
    The loss function is the sum of a bias term, and a term that quantifies how well we can separate the data into subsets.
    From Eq.~(\ref{eq-GradFull}), it is pretty clear that the value of $\Delta p_i$ controls which term do we try to minimize faster during gradient descent. 
    Thus, it is a free parameter, and it is conceivable that in different problems and different stages of the descent, we may want to tune it to achieve better results.
    \item {\bf Tree-Boosting.}
    Ensemble methods with boosted trees, such as XGBoost \cite{Fri1999, CheGue2016}, are proven to be one of the most reliable and efficient technique for many supervised-learning problems.
    For a causal problem, we can build the first tree as the tree-based model in \cite{RzeJar2012}.
    Starting from the first tree, if we take the analogy between our framework and a regression problem seriously, it seems like we can take $(-\frac{\partial L}{\partial p_i})$ defined in Eq.~(\ref{eq-GradFull}) as the target value for boosting.
    How well does that strategy work after many boosting rounds is a very interesting question that requires further study.
\end{enumerate}
The list here is certainly not exhaustive.
Basically, any advance techniques or obstacles we can encounter in standard gradient descent has an exact correspondence here, and may require extra care.
We look forward to seeing more development in this field and further enlarge our bag of tools in dealing with the large variety of causal machine-learning problems.

\acknowledgments

We thank Bertrand Brelier and Eugene Chen for interesting discussions.
In particular, Dr. Brelier introduced us to the field of causal machine learning and provided a lot of guidance to background materials.
Computation in this work is done on free cloud resources provided by Databricks Community Edition (https://community.cloud.databricks.com/).

\end{document}